\documentclass[conference]{IEEEtran}
\IEEEoverridecommandlockouts
\usepackage{cite}
\usepackage{marvosym}
\usepackage{amsmath,amssymb,amsfonts}
\usepackage{algorithmic}
\usepackage{graphicx}
\usepackage{textcomp}
\usepackage[table,xcdraw]{xcolor}
\usepackage{multirow} %zjh
\usepackage{subfig} %zjh
\usepackage[numbers]{natbib} %zjh
\usepackage{multirow}
\usepackage{booktabs}
\def\BibTeX{{\rm B\kern-.05em{\sc i\kern-.025em b}\kern-.08em
    T\kern-.1667em\lower.7ex\hbox{E}\kern-.125emX}}

\begin{document}

\title{Triple Path Enhanced Neural Architecture Search for Multimodal Fake News Detection \\
}

\author{
% \begin{center}
% \centering
Bo Xu$^{1{\dagger}}$, Qiujie Xie$^{2{\dagger}}$, Jiahui Zhou$^3$, Linlin Zong$^{3*}$  \\
{\small $^1$School of Computer Science and Technology, Dalian
University of Technology} 
{\small $^2$School of Computer Science, Fudan University} \\
{\small $^3$School of Software, Dalian University of Technology} \\
% {$^4$Key Laboratory for Ubiquitous Network and Service Software of Liaoning Province, School of Software,\\ Dalian University of Technology} \\
\texttt{\small \{llzong, xubo\}@dlut.edu.cn, qjxie22@m.fudan.edu.cn, zjhjixiang@mail.dlut.edu.cn}
% \end{center}

\thanks{$^{\dagger}$ Equal contribution. $^{*}$ Corresponding author.}
\thanks{This work was supported in part by the Fundamental Research Funds for the Central Universities (DUT24MS003), and the Liaoning Provincial Natural Science Foundation Joint Fund Program(2023-MSBA-003).}
}
% \IEEEoverridecommandlockouts % 启用自定义命令

% \author{\IEEEauthorblockN{1\textsuperscript{st} Linlin Zong}
% \IEEEauthorblockA{\textit{School of Software} \\
% \textit{Dalian University of Technology}\\
% llzong@dlut.edu.cn}
% \and
% \IEEEauthorblockN{2\textsuperscript{nd} Qiujie Xie}
% \IEEEauthorblockA{\textit{School of Computer Science} \\
% \textit{Fudan University}\\
% qjxie22@m.fudan.edu.cn}
% \and
% \IEEEauthorblockN{3\textsuperscript{rd} Jiahui Zhou}
% \IEEEauthorblockA{\textit{School of Software} \\
% \textit{Dalian University of Technology}\\
% zjhjixiang@mail.dlut.edu.cn}
% \and
% \IEEEauthorblockN{4\textsuperscript{th} Bo Xu}
% \IEEEauthorblockA{\textit{School of Computer Science and Technology} \\
% \textit{Dalian University of Technology}\\
% xubo@dlut.edu.cn}

% \and
% \and

\maketitle
% \IEEEaftertitletext{\vspace{-2cm}} % 调整标题和正文之间的间距
% \vspace{-1cm}

\begin{abstract}
Multimodal fake news detection has become one of the most crucial issues on social media platforms. Although existing methods have achieved advanced performance, two main challenges persist: (1) Under-performed multimodal news information fusion due to model architecture solidification, and (2) weak generalization ability on partial-modality contained fake news. To meet these challenges, we propose a novel and flexible triple path enhanced neural architecture search model MUSE. MUSE includes two dynamic paths for detecting partial-modality contained fake news and a static path for exploiting potential multimodal correlations. Experimental results show that MUSE achieves stable performance improvement over the baselines.
\end{abstract}


\section{Introduction}
The emergence of social media has revolutionized information sharing, leading to an alarming surge in deliberately fabricated 
% or misleading 
stories known as fake news \cite{DBLP:conf/icmcs/ZengWLLHS23}. This trend has been further exacerbated by the advent of large language models~\cite{ren2021comprehensive, DBLP:journals/corr/abs-2303-08774, DBLP:conf/acl/ZongZLL0024}
%, such as ChatGPT \cite{DBLP:journals/corr/abs-2303-08774} and Gemini \cite{DBLP:journals/corr/abs-2312-11805}, 
which can generate a substantial volume of deceptive articles. Fake news often involves various modalities (text, images, and videos etc.), requiring the integration of multimodal information for more accurate detection. However, each modality contains distinct information needing specific analyses, like visual cues in images and linguistic indicators in text. Hence, effectively detecting multimodal fake news on social media 
% platforms 
remains a critical pursuit for researchers.

% % Addressing the proliferation of fake news is critical in purifying the online environment. 
% Fake news often involves various modalities (text, images, and videos etc.), requiring the integration of multimodal information for more accurate detection. However, each modality contains distinct information needing specific analyses, like visual cues in images and linguistic indicators in text. Hence, effectively detecting multimodal fake news on social media 
% % platforms 
% remains a critical pursuit for researchers.

% Addressing the proliferation of fake news is critical in purifying the online environment. Fake news often involves various data types, including text, images, and videos et al., necessitating the integration of multimodal information for more accurate detection. However, detecting multimodal fake news poses a unique challenge as each modality contains distinct information needing specific analyses. For instance, images might carry visual cues while textual content may reveal linguistic indicators of fake news. Hence, effectively detecting multimodal fake news on social media platforms remains a critical pursuit for researchers.

\begin{figure}[h]
  \centering
  \setlength{\abovecaptionskip}{0.cm}
  \includegraphics[width=0.85\linewidth]{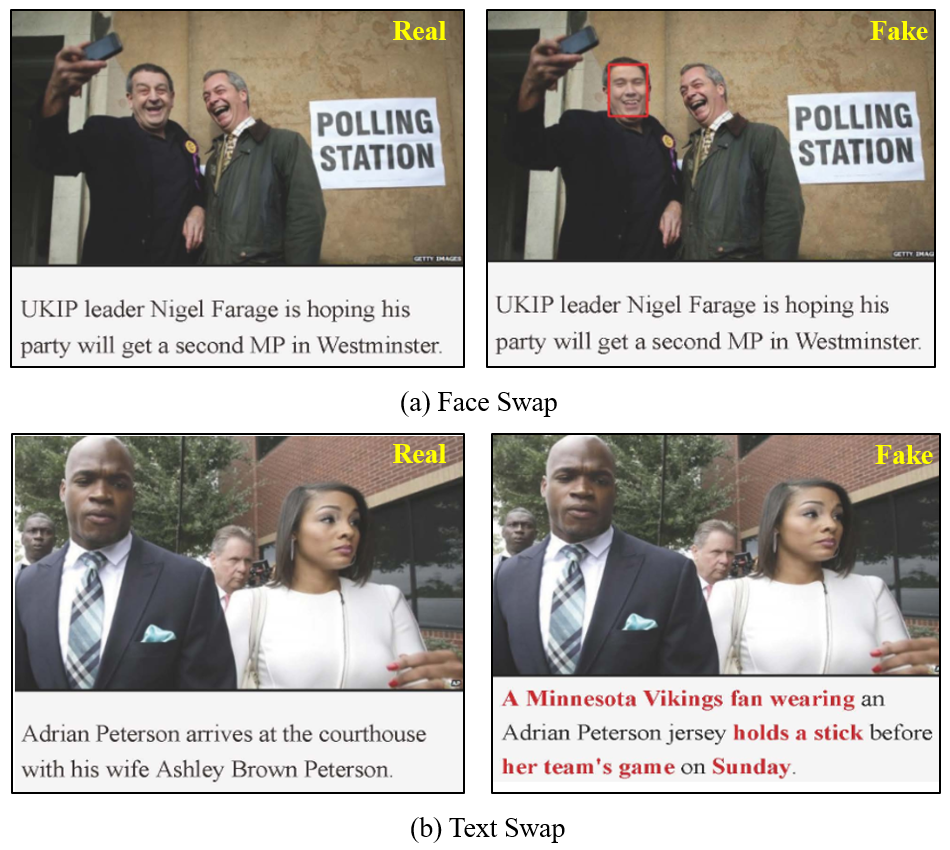}
  \caption{Example for various types of fake news.}
  \label{fig:example}
 \vspace{-1.5em}
\end{figure}

\begin{figure*}[htp]
	\centering
	\includegraphics[width=0.85\linewidth]{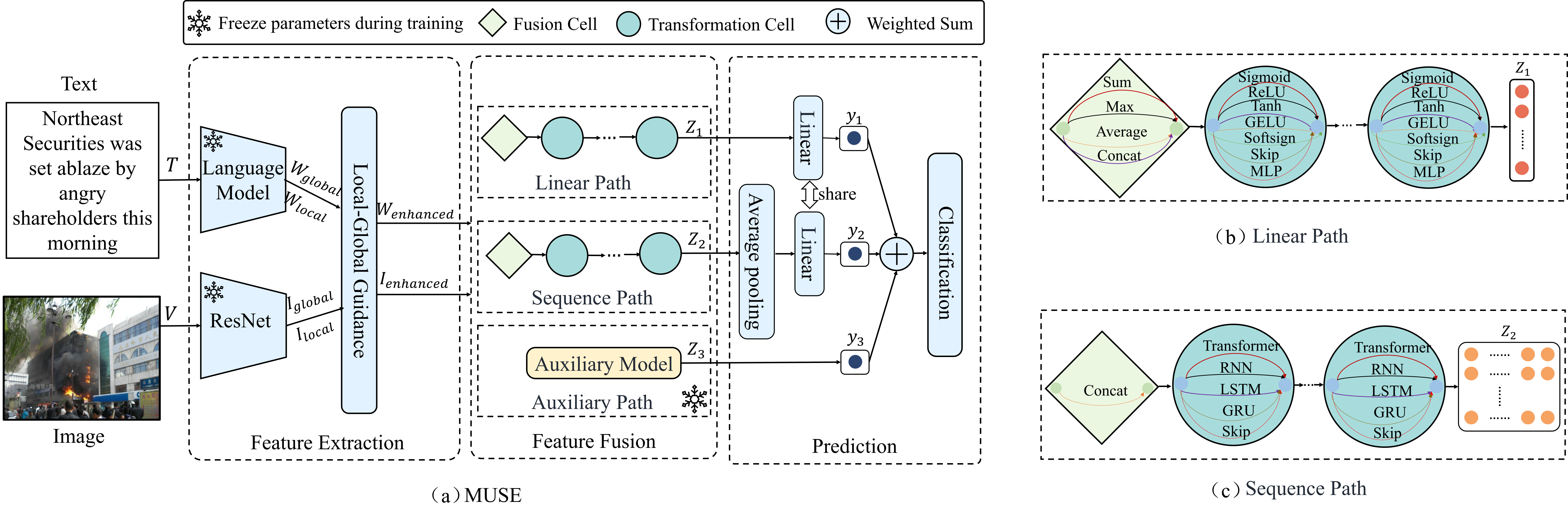}
	\caption{Overall architecture of the MUSE model.}
	\label{fig:architecture_MUSE} 
   \vspace{-1.5em}
\end{figure*}

% While substantial advancements have been achieved in multimodal false news detection %\cite{wang2018eann,qian2021hierarchical,zhou2020mathsf,DBLP:conf/icmcs/ZhouYYQZ23}, 
Existing fake news detection methods focus on learning effective semantic representations. For example, the attention mechanism \cite{chen2018call,qian2021hierarchical}, modality consistency \cite{zhou2020mathsf,xue2021detecting}, contrastive pretraining \cite{DBLP:conf/icmcs/ZhouYYQZ23} were used to handle the complicated interactions of multimodal data. In addition, Wang et al. \cite{wang2018eann} introduced event classification; 
% as an auxiliary task for fake news detection; 
Qi et al. \cite{qi2021improving} proposed a novel framework using cross-modal correlations.
While substantial advancements have been achieved in multimodal false news detection, practical challenges persist, necessitating further optimization of detection models.
Firstly, diverse categories of fake news exhibit semantic conflicts and content distortions, as shown in \figurename~\ref{fig:example}. The integration of complementary information from different modalities remains an open question, as current models often rely on fixed architectures.
Secondly, in real-world scenarios, multimodal news may involve only a subset of available modalities, making it difficult for models trained on partial-modality to adapt to dynamically generated content. 
% Despite promising performances of existing methods, effectively detecting partial-modality fake news remains challenging due to limited generalization capabilities.
% While substantial advancements have been achieved in multimodal false news detection \cite{wang2018eann,qian2021hierarchical,zhou2020mathsf,DBLP:conf/icmcs/ZhouYYQZ23}, practical challenges persist, underscoring the need for further optimization of multimodal false news detection models.
% Firstly, diverse categories of fake news exhibit semantic conflicts and content distortions, exemplified by \figurename~\ref{fig:example}. 
% %Tailored methods are essential to fuse effective features from diverse modalities.%
% The effective integration of complementary information from diverse modalities remains an open question. Present detection models typically rely on fixed architectures, constraining their ability to capture diverse semantic conflicts and content distortions present in fake news.
% Secondly, in real-world scenarios, multimodal news often involves only a subset of available modalities. Some news items may incorporate both images and text, while others rely solely on a single modality. Models trained on partial-modality fake news struggle to adapt to dynamically generated multimodal news. Consequently, the models' capacity to generalize becomes crucial in detecting fake news. Despite promising performances of existing methods, effectively detecting partial-modality fake news remains challenging due to limited generalization capabilities.

To address these challenges, we propose the MUltimodal neural architecture Search Enhanced fake news detection model~(MUSE). MUSE leverages Neural Architecture Search (NAS) techniques \cite{DBLP:conf/iclr/LiuSY19, chang2024survey,zheng2023can} to enhance the model's generalization capacity. Specifically, MUSE incorporates two dynamic paths and one static path, offering flexibility in dynamically modifying its network architecture to identify the most effective operators for extracting multimodal fusion features. Our contributions can be summarized as follows: 

\begin{itemize}
    \item MUSE leverages NAS techniques to dynamically adjust network architecture for improved feature fusion and generalization capacity.    
    \item MUSE can adapt to diverse types of fake news. It consistently demonstrates superior performance compared to baselines, particularly for data lacking in modalities.
\end{itemize}

\section{Method}
\label{sec:method}

\subsection{General Framework}
We focus on fake news detection involving text and image modalities. In a dataset $D = \left\{ X_{i},Y_{i} \right\}_{i = 1}^{N}$ containing $N$ news samples and corresponding labels, the sample $X{i}$ typically includes a textual segment $T$ and an image $V$. To mirror real-world scenarios, we acknowledge the presence of incomplete data within $D$. Samples lack text are indicated as $X_{i} = \left\{ {\bar{T},V} \right\}$, while others lack images are expressed as $X_{i} = \left\{ {T,\bar{V}} \right\}$. Given this context, the research problem entails identifying the mapping
\begin{math}
	f = \left\{ {T/\bar{T},V/\bar{V}} \right\}~\overset{\theta}{\rightarrow}~Y \in \left\{ 0,1 \right\}
\end{math},
where $\theta$ denotes the model parameters, and $Y = 1$ and $Y = 0$ denote labels for fake and authentic news, respectively.

Given the diverse fake news types and the possibility of missing data, we introduce the MUSE model (\figurename~\ref{fig:architecture_MUSE}(a)). MUSE contains three stages for fake news detection: feature extraction, feature fusion, and news prediction. Details are described in the following sections.
%Finally, a linear layer serves as the classifier in the prediction stage to detect fake news.

\subsection{Multimodal Feature Extraction}
\label{subsec:feature_extraction}

We use the pre-trained language model to extract the text feature matrix, $W_{local} = \left\{ w_{1};\ldots ;w_{K_{T}} \right\} \in \mathbb{R}^{K_{T} \times D_{T}}$, where $K_{T}$ is the number of words in $T$, and $D_{T}$ is the dimension of feature vectors. Similarly, we use the pre-trained ResNet-50 \cite{he2016deep} to extract the regional features of a given image $V$, namely $I_{local} = $ResNet$(V) = \left\{ {i_{1};\ldots i_{K_{V}}} \right\} \in \mathbb{R}^{K_{V}\times D_{V}}$, where $K_{V}$ is the number of regions in the image, and $D_{V}$ is the dimension of the feature vectors.

Since the raw features may contain inadequate information that condenses the context and high-level semantics, we adopt the Global-Local Guidance \cite{qu2021dynamic} to further refine the multimodal features. 
% Specifically, we use the raw features of one modality as a guidance to fine-grain the features of another modality and vice versa, thereby enhancing the modality representations. 
Taking the text features as an example, we use \eqref{eq4} and \eqref{eq5} to obtain the enhanced text feature $W_{enhanced}$, where $I_{global} \in \mathbb{R}^{D_{V}}$ is the fine-grained feature by applying the average pooling to $I_{local}$, $\odot$ represents the Hadamard product, $FC$ represents the fully connected layer, Norm$(\cdot )$ represents L2-normalization.
\vspace{-0.8em}
\begin{equation}\label{eq4}
\small
	d = W_{local}~\odot~FC\left( I_{global} \right)
 \vspace{-0.5em}
\end{equation}
\begin{equation}\label{eq5}
\small
	W_{enhanced} = \left( {1 + Norm(d)} \right)~\odot{~W}_{local}
\vspace{-0.5em}
\end{equation}

Given that some news samples only contain partial modalities, we use \eqref{eq6} to calculate the $W_{enhanced}$ matrix of these samples. We obtain the enhanced image features $I_{enhanced} $ in the same way.
\vspace{-0.8em}
\begin{equation}\label{eq6}
\small
	{W}_{enhanced} = \left\{ \begin{matrix}
		{0~when~input =}
		{\left\{ {{\bar{T}},V} \right\},}\\
		{W_{local}~when~input =} \{{ T,{\bar{V}}}\}\\
	\end{matrix} \right.
\end{equation}

\begin{table*}
	\center
    \setlength{\abovecaptionskip}{2pt}
    \renewcommand{\arraystretch}{1.1}
	\scriptsize
	\caption{Comparison Results on the WEIBO and PHEME Datasets.}
	\label{tab:comparePHEME}
    \setlength\tabcolsep{5pt} %调整表格列间距
    \resizebox{0.83\textwidth}{!}{
	\begin{tabular}{c|c|ccc|ccc|c|ccc|ccc}
		\toprule
		\multirow{3}{*}{Model} & \multicolumn{7}{c|}{WEIBO}& \multicolumn{7}{c}{PHEME}\\  \cline{2-15}
		&  \multirow{2}{*}{Acc.} & \multicolumn{3}{c|}{Fake News} & \multicolumn{3}{c|}{Real News} &  \multirow{2}{*}{Acc.} & \multicolumn{3}{c|}{Fake News} & \multicolumn{3}{c}{Real News}\\ \cline{3-8}\cline{10-15}
		&                           & Precision   & Recall   & F1    & Prec.   & Recall   & F1  &  & Precision   & Recall   & F1    & Precision   & Recall   & F1   \\ \midrule
		GRU & 0.702&0.671&0.794&0.727&0.747&0.609&0.671& 0.832&0.782&0.712&0.745&0.855&0.896&0.865\\
		CAMI & 0.740&0.736&0.756&0.744&0.747&0.723&0.735& 0.779&0.732&0.606&0.663&0.799&0.875&0.835\\
		% SVM-TS & 0.640&0.741&0.573&0.646&0.651&0.798&0.711& 0.639&0.546&0.576&0.560&0.729&0.705&0.717\\
		TextGCN & 0.787&0.975&0.573&0.727&0.712&0.985&0.827& 0.828&0.775&0.735&0.737&0.827&0.828&0.828\\
		EANN & 0.782&0.827&0.697&0.756&0.752&0.863&0.804& 0.681&0.685&0.664&0.694&0.701&0.750&0.747\\
		MVAE & 0.824&0.854&0.769&0.809&0.802&0.875&0.837& 0.852&0.806&0.719&0.760&0.871&0.917&0.893\\
		% SAFE & 0.763&0.833&0.659&0.736&0.717&0.868&0.785& 0.811&0.827&0.559&0.667&0.806&0.940&0.866\\
		att\_RNN &0.772&0.854&0.656&0.742&0.720&0.889&0.795&0.850&0.791&0.749&0.770&0.876&0.899&0.888\\
		%&SpotFake &0.892&0.902&0.964&\textbf{0.932}&0.847&0.656&0.739\\
		SpotFake* &0.869&0.877&0.859&0.868&0.861&0.879&0.870&0.823&0.743&0.745&0.744&0.864&0.863&0.863\\
		SpotFake+ &0.870&0.887&0.849&0.868&0.855&0.892&0.873&0.800&0.730&0.668&0.697&0.832&0.869&0.850\\
		HMCAN & 0.885&0.920&0.845&0.881&0.856&\textbf{0.926}&0.890& 0.871&\textbf{0.860}&0.781&0.818&0.876&\textbf{0.924}&0.900\\
		MCAN & 0.899&\textbf{0.913}&0.889&0.901&0.884&0.909&0.897&-&-&-&-&-&-&-\\
		CAFE & 0.840&0.855&0.830&0.842&0.825&0.851&0.837&-&-&-&-&-&-&-\\
		FND-CLIP & 0.907& 0.914& 0.901& 0.908& 0.914& 0.901& 0.907&-&-&-&-&-&-&-\\
		\midrule
        \rowcolor[HTML]{EFEFEF}
		MUSE & \emph{0.905}&0.898&0.915&\emph{0.906}&0.913&0.896&\emph{0.904}& 0.871&\textbf{0.860}&0.781&0.818&0.876&\textbf{0.924}&0.900\\
        \rowcolor[HTML]{EFEFEF}
		MUSE-discrete & \textbf{0.914}&0.905&\textbf{0.926}& \textbf{0.916}&\textbf{0.924}&0.903& \textbf{0.913}& \textbf{0.878}&0.836&\textbf{0.838}&\textbf{0.837}&\textbf{0.902}&0.902&\textbf{0.902}\\
  \bottomrule
	\end{tabular}
 }
 \vspace{-2.0em}
\end{table*}

\subsection{Multimodal Feature Fusion}
In this stage, we design two dynamic paths and one static path. The dynamic paths, using DARTS \cite{DBLP:conf/iclr/LiuSY19}, to address the rigidity of model structure. Meanwhile, the static path preserves essential modality correlations throughout training. 
\subsubsection{Dynamic Path in MUSE}
 We optimize architecture with DARTS by adapting it from a multi-input-output graph to a single-input and single-output structure. This change ensures accuracy in detecting fake news, while also reducing the time required for architecture search.
 % We use DARTS for architecture optimization. To achieve this, we modify DARTS by switching it from a multi-input-output directed acyclic graph structure to a single-input and single-output linked structure. 
 %This change ensures accuracy in detecting fake news, while also reducing the time required for architecture search.

In our modified approach, each dynamic path comprises a linked structure with $n$ cells. Each cell, serving as the building block, searches for the most effective neural network for input data. These cells are categorized into fusion cells and transformation cells based on their search targets. Fusion cells take in enhanced sample features and search for the optimal architecture to fuse these features, while transformation cells search for deep neural architectures to uncover hidden semantic information within the fused features.

Mathematically, each intermediate cell $h^{(j)}$ represents a potential feature representation and connects to preceding cell $h^{(i)}, i<j$ through directed operator edges $e^{(i,j)}$. Therefore, each intermediate cell can be represented using \eqref{eq1}.
\vspace{-0.5em}
\begin{equation}\label{eq1}
\small
	h^{(j)} = \sum_{i<j}e^{({i,j})} h^{(i)}
\vspace{-0.5em}
\end{equation}
To create a continuous search space, each edge $e^{(i,j)}$ is associated with a weight $\alpha_{o}^{(i,j)}$ for the operator $o$. The mixed operator applied to the feature representation $h$ is represented by \eqref{eq2}.
\vspace{-0.5em}
\begin{equation}\label{eq2}
\small
	{\hat{o}}^{({i,j})}(h) = {\sum\limits_{o \in O}{\frac{\exp\left( \alpha_{o}^{({i,j})} \right)}{\sum\limits_{o^{'} \in O}{\exp\left( \alpha_{o^{'}}^{({i,j})} \right)}}o(h)}}
\vspace{-0.5em}
\end{equation}

where $O$ represents the set of candidate operators. Thus, the task simplifies to optimizing a set of continuous variables $\alpha_{o}^{(i,j)}$.
After completing the architecture search, we select the most effective operator $o^{(i,j)}$ with the highest weight on each directed edge $e^{(i,j)}$ and discard other operators to obtain a discrete model, as demonstrated in \eqref{eq3}.
\vspace{-0.5em}
\begin{equation}\label{eq3}
\small
	o^{(i,j)} = {argmax}_{o \in O}\alpha_{o}^{(i,j)}
 \vspace{-0.5em}
\end{equation}

(1) Linear path. Considering different operators can effectively processing certain types of news, we design a linear operator search path to search for an effective linear neural network operator for one-dimensional feature representations. As shown in \figurename~\ref{fig:architecture_MUSE}(b), the linear path consists of one fusion cell and $n-1$ transformation cells. The search space of the fusion cell includes four operators: \textit{Sum}, \textit{Max}, \textit{Average} and \textit{Concat}.
%, which can be extended for different tasks.
% From the perspective of operations, a neural network can be considered as a deep learning operator applied on input data. \textbf{Since different operators require different dimensionalities of data, the corresponding cell construction processes of operators are different.} For example, when the search space of the fusion cell contains the \textit{Average} and the \textit{Sum} operators, we use the linear layer to map the enhanced features of each modality to the same common space; When the search space contains the  \textit{Concat} operator, we use the linear layer to map the output features to the same space $s$ to ensure that the output dimensions of different fusion units are consistent. 
Taking the \textit{Sum} operator as an example, the fused feature $X_{fused}$ is calculated as follows:
\vspace{-0.5em}
\begin{equation}\label{eq7}
\small
	X_{fused} = Sum\left( FC(W_{enhanced}), FC(I_{enhanced}) \right)
 \vspace{-0.8em}
\end{equation}
% After obtaining the fused feature $X_{fused}$, $n-1$ transformation cells perform deep transformation to mine the hidden information of the fused feature. %As shown in Fig. \ref{fig:linear},
The search space of the transformation cell is composed of five commonly used activation functions, namely \textit{Sigmoid}, \textit{ReLU}, \textit{Tanh}, \textit{GELU}, \textit{Softsign}, together with the \textit{Skip} and the \textit{Multi-Layer Perception(MLP}) operator. Taking the \textit{MLP} operator as an example, the outputs can be calculated as follows. 
% Taking the \textit{Skip} operator and the \textit{MLP} operator as examples, the outputs can be calculated as follows.
After $n-1$ transformation cells, we obtain the output vector $Z_{1}$ of the linear path.
\vspace{-0.5em}
\begin{equation}\label{eq11}
\small
	X_{fused}^{'} = \left\{ \begin{matrix}
		{Skip\left( X_{fused} \right)}\\
            ...\\
		{MLP\left( X_{fused} \right)}\\
	\end{matrix} \right.
    \vspace{-0.5em}
\end{equation}

% \begin{equation}
% \small
%     Skip\left( X_{fused} \right) = X_{fused}
%     \vspace{-2em}
% \end{equation}

\begin{equation}
\small
    MLP\left( X_{fused} \right) = FC\left( Sigmoid\left( FC\left( X_{fused} \right) \right) \right)
    \vspace{-0.5em}
\end{equation}
% \vspace{-1em}
% \begin{equation}\label{eq11}
% \small
% 	X_{fused}^{'} = \left\{ \begin{matrix}
% 		{Skip\left( X_{fused} \right)}\\
% 		{MLP\left( X_{fused} \right)}\\
% 	\end{matrix} \right.
%     \vspace{-1.5em}
% \end{equation}

% \begin{equation}
% \small
%     Skip\left( X_{fused} \right) = X_{fused}
%     \vspace{-2em}
% \end{equation}

% \begin{equation}
% \small
%     MLP\left( X_{fused} \right) = FC\left( Sigmoid\left( FC\left( X_{fused} \right) \right) \right)
%     \vspace{-1em}
% \end{equation}

(2) Sequence path. To model multi-dimensional sequence features with richer inter-sequence information, we design a sequence path to search for complex architectures embedded with hidden news relationships. The sequence path consists of one fusion cell and $m-1$ transformation cells. As shown in \figurename~\ref{fig:architecture_MUSE}(c), the search space of the fusion cell consists of the multi-dimensional concatenation operator \textit{Concat}, and the fused features $X_{fused}$ after fusion cell is computed as follows.
\vspace{-0.5em}
\begin{equation}\label{eq12}
\small
	X_{fused} = Concat\left( FC\left( W_{enhanced} \right),FC\left( I_{enhanced} \right) \right)
 \vspace{-0.5em}
\end{equation}
%When the enhanced features of text modality are missing, we remove the influence of the missing data ${\bar{W}}_{enhanced}$ and make $I_{enhanced}$ completely dominates $X_{fused}$.%, as shown in Eq. (\ref{eq13}). Similar ways are used for news with missing images.
%\begin{equation}\label{eq13}
%	\begin{split}
%		X_{fused} & = FC(I\_ enhanced~)~~\\
%		%when~input & = \left\{ {\bar{W}}~\overline{}\_ enhanced,I\_ enhanced~ \right\}
%	\end{split}
%\end{equation}
For the transformation cell, %in the sequence path its search space needs deep operators with strong feature mining ability. Therefore, 
we use the \textit{Transformer}, \textit{RNN}, \textit{LSTM}, \textit{GRU} and the \textit{Skip} operator as the search space. After $m-1$ transformation cells, we name the output vector of the sequence path $Z_2$.

\subsubsection{Static Path in MUSE}
%To extract valuable yet unseen auxiliary information, 
We incorporate a static path aimed at capturing latent details for detection, producing outputs $Z_3$ that represent the auxiliary information. 
%In contrast to the dynamic path, 
This auxiliary path remains constant throughout the training process. 
In practical applications, this path can be predefined and optimized using non-parameter strategies tailored to different task objectives. For example, 
in datasets with quality concerns, such as the PHEME dataset\cite{zubiaga2017exploiting}, 
% where samples are purposefully gathered around five social events, 
we address the issue of sample reuse 
% to generate adequately similar samples. We 
by proposing a sample reference strategy in the static path. Here, samples are clustered without supervision, and the average prediction score of similar samples within the same batch serves as the outputs $Z_3$.
We provide two ways to build the static path.
For datasets characterized by %high-quality and 
complete multimodal relationships like the WEIBO dataset \cite{jin2017multimodal}, we design a similarity Siamese Network as the auxiliary path. 
%This network comprises two identical subnets.
In datasets with quality concerns, such as the PHEME dataset \cite{zubiaga2017exploiting}, where samples are purposefully gathered around five social events, we address the issue of sample reuse by generating adequately similar samples. 
% We propose a sample reference strategy to eliminate the need for the auxiliary model. 
Here, samples are clustered without supervision, and the average prediction score of similar samples within the same batch serves as the auxiliary information.

\vspace{-0.5em}
\begin{table}[htp]
% \vspace{-0.8em}
	\center
    \renewcommand{\arraystretch}{1.1}
    \setlength{\abovecaptionskip}{0.cm}
    \setlength\tabcolsep{4.5pt}
	\scriptsize
	\caption{Results on the WEIBO\_Partial Dataset.}
	\label{tab:partial_experiment}
	\begin{tabular}{c|c|ccc|ccc}
		\toprule
		\multirow{2}{*}{Model} & \multirow{2}{*}{Acc.} & \multicolumn{3}{c|}{Fake News} & \multicolumn{3}{c}{Real News}  \\ \cline{3-8}
		&                           & Precision   & Recall   & F1    & Precision   & Recall   & F1     \\ \hline
		MCAN                   & 0.678                      & 0.635        & 0.607     & 0.621  & 0.696        & 0.721     & 0.708      \\
		Baseline               & 0.690                      & 0.655        & 0.640     & 0.648  & 0.717        & 0.730     & 0.723  \\ \midrule
        \rowcolor[HTML]{EFEFEF}
		MUSE                   & \textbf{0.750}                      & 0.701        & 0.764     & \textbf{0.731}  & 0.796        & 0.734     & \textbf{0.767}                \\ \bottomrule
	\end{tabular}
 \vspace{-1.5em}
\end{table}

\subsection{Fake News Prediction}
We combine the output vectors $Z_1$, $Z_2$, and $Z_3$ from three different paths using a linear layer as the final classifier to fuse their classification vectors $y_1$, $y_2$, and $y_3$, 
%By dynamically assigning weights to $y_1$, $y_2$, and $y_3$, we compute the final prediction score $y_i$ for any news sample $X_i$
as shown in \eqref{eq15}.
\vspace{-0.7em}
\begin{equation}\label{eq15}
\small
	y_{i} = \beta*Scale\left( y_{1} \right) + \gamma*Scale\left( y_{2} \right) + \delta*Scale\left( y_{3} \right)
 \vspace{-0.7em}
\end{equation}
where $\beta$, $\gamma$ and $\delta$ are learnable variables. 
The function Scale$(\cdot)$ scales the classification vector values of the three paths to a uniform interval. 
%We employ a dynamic weighting strategy to guide the model in assigning path weights based on the difficulty levels of detecting news.
 %For example, when processing news with high data quality, it is preferable to use low-complexity operators for the linear path. Thus, the model will automatically increase the weight $\beta$ in the linear path. On the contrary, when dealing with low-quality news with missing modalities, complex operators will be adopted for the sequence path so as to promote the model to increase the weight $\gamma$ in the sequence path.
Finally, we use the binary cross entropy loss function as the objective function shown in \eqref{eq16}.
\vspace{-0.5em}
\begin{equation}\label{eq16}
\small
	{Loss}_{i} = - \left( Y_{i}*{\log\left( y_{i} \right)} + \left( 1 - Y_{i} \right)log\left( 1 - y_{i} \right) \right)
 \vspace{-0.5em}
\end{equation}

where $Y_i$ is the true label of the sample $X_i$, and $y_i$ is the prediction probability of the sample $X_i$.

\begin{table*}[htp]\small
\vspace{-0.8em}
\center
\renewcommand{\arraystretch}{1.1}
    \caption{Results of Ablation Experiment on the WEIBO and PHEME Datasets for Operators.}
    \label{tab:ablation_oper}
    \setlength{\abovecaptionskip}{0.cm}
	\scriptsize
    \setlength\tabcolsep{5pt}
    \resizebox{0.65\textwidth}{!}{
    \begin{tabular}{c|c|c|ccc|ccc}
    \toprule
    %Dataset                & Number of Operators&$Accuracy$&$Precision_{f}$&$Recall_{f}$&$F1_{f}$&$Precision_{r}$&$Recall_{r}$&$F1_{r}$
    \multirow{2}{*}{Dataset}&		\multirow{2}{*}{Number of Operators} & \multirow{2}{*}{Accuracy} & \multicolumn{3}{c|}{Fake News} & \multicolumn{3}{c}{Real News} \\ \cline{4-9}
		&       &                    & Precision   & Recall   & F1    & Precision   & Recall   & F1     \\ \hline
    \multirow{6}{*}{WEIBO}  &All Operators & 0.904&0.875&0.943&0.908&0.938&0.865&0.900\\
                            &6 Operators & 0.900&0.918&0.880&0.898&0.884&0.921&0.902\\
                            &5 Operators & 0.904&0.911&0.897&0.904&0.898&0.912&0.905\\
                            &4 Operators & 0.903&0.910&0.896&0.903&0.897&0.911&0.904\\
                            &3 Operators & 0.911&0.913&0.909&0.911&0.909&0.913&0.911\\
                            &2 Operators & 0.911&0.890&0.939&0.914&0.935&0.884&0.909\\
                            &1 Operators & \textbf{0.914}&0.905&0.926&0.916&0.924&0.903&0.913\\ \hline
    \multirow{6}{*}{PHEME}  &All Operators & 0.866&0.865&0.758&0.808&0.866&0.930&0.897\\
                            &6 Operators & 0.868&0.877&0.857&0.867&0.860&0.880&0.870\\
                            &5 Operators & 0.871&0.870&0.769&0.816&0.871&0.931&0.900\\
                            &4 Operators & 0.868&0.843&0.795&0.818&0.882&0.912&0.897\\
                            &3 Operators & 0.872&0.842&0.811&0.826&0.900&0.909&0.899\\
                            &2 Operators & \textbf{0.878}&0.836&0.838&0.837&0.902&0.902&0.902\\
                            &1 Operators & 0.847&0.817&0.762&0.789&0.864&0.898&0.881\\ \bottomrule
    \end{tabular}
}
    \vspace{-0.5em}
\end{table*}

\begin{table}[htp]\small
\vspace{-0.8em}
	\center
    \setlength\tabcolsep{4pt}
    \renewcommand{\arraystretch}{1.1}
    \setlength{\abovecaptionskip}{0.cm}
	\caption{Ablation study of paths on WEIBO.}
	\scriptsize
	\label{tab:ablation_path}
	\begin{tabular}{c|c|ccc|ccc}
		\toprule
		%Dataset                & Experimental Design&$Accuracy$&$Precision_{f}$&$Recall_{f}$&$F1_{f}$&$Precision_{r}$&$Recall_{r}$&$F1_{r}$
		%\hline
		\multirow{3}{*}{Ablation} & \multirow{2}{*}{Acc.} & \multicolumn{3}{c|}{Fake News} & \multicolumn{3}{c}{Real News} \\ \cline{3-8}
		&  & Precision   & Recall   & F1    & Precision   & Recall   & F1     \\ \hline
		% \multicolumn{8}{|c|}{Weibo}  \\ \hline
		w/o Linear& 0.847&0.914&0.766&0.834&0.798&0.928&0.858\\
		w/o Sequence& 0.894&0.886&0.906&0.896&0.903&0.883&0.893\\
		w/o Auxiliary & 0.844&0.916&0.759&0.830&0.794&0.930&0.856\\ %\cline{2-8}
        \midrule
        \rowcolor[HTML]{EFEFEF}
		MUSE & \textbf{0.905}&0.898&0.915&\textbf{0.906}&0.913&0.896&\textbf{0.904}\\ \bottomrule
		% \multicolumn{8}{|c|}{PHEME}  \\\hline
		% w/o Linear& 0.870&0.828&0.820&0.824&0.893&0.898&0.896\\
		% w/o Sequence& 0.855&0.807&0.804&0.806&0.883&0.886&0.885\\
		% w/o Auxiliary& 0.867&0.836&0.801&0.818&0.885&0.907&0.895\\ %\cline{2-8}
		% MUSE & \textbf{0.871}&0.860&0.781&\textbf{0.818}&0.876&0.924&\textbf{0.900}\\ \hline
	\end{tabular}
 % \vspace{-2em}
\end{table}

\begin{table}[htp]\small
\vspace{-0.8em}
	\center
    \setlength\tabcolsep{4pt}
    \renewcommand{\arraystretch}{1.1}
	\caption{Ablation study of paths on PHEME.}
    \setlength{\abovecaptionskip}{0.cm}
	\scriptsize
	\label{tab:ablation_path_pheme}
	\begin{tabular}{c|c|ccc|ccc}
		\toprule
		%Dataset                & Experimental Design&$Accuracy$&$Precision_{f}$&$Recall_{f}$&$F1_{f}$&$Precision_{r}$&$Recall_{r}$&$F1_{r}$
		%\hline
		\multirow{3}{*}{Ablation} & \multirow{2}{*}{Acc.} & \multicolumn{3}{c|}{Fake News} & \multicolumn{3}{c}{Real News} \\ \cline{3-8}
		&  & Precision   & Recall   & F1    & Precision   & Recall   & F1     \\ \hline
		%\multicolumn{8}{|c|}{Weibo}  \\ \hline
		w/o Linear& 0.870&0.828&0.820&0.824&0.893&0.898&0.896\\
		w/o Sequence& 0.855&0.807&0.804&0.806&0.883&0.886&0.885\\
		w/o Auxiliary & 0.867&0.836&0.801&0.818&0.885&0.907&0.895\\ %\cline{2-8}
        \midrule
        \rowcolor[HTML]{EFEFEF}
		MUSE &\textbf{0.871}&0.860&0.781&\textbf{0.818}&0.876&0.924&\textbf{0.900}\\ \bottomrule
%		\multicolumn{8}{|c|}{PHEME}  \\\hline
%		w/o Linear& 0.870&0.828&0.820&0.824&0.893&0.898&0.896\\
%		w/o Sequence& 0.855&0.807&0.804&0.806&0.883&0.886&0.885\\
%		w/o Auxiliary& 0.867&0.836&0.801&0.818&0.885&0.907&0.895\\ %\cline{2-8}
%		MUSE & \textbf{0.871}&0.860&0.781&\textbf{0.818}&0.876&0.924&\textbf{0.900}\\ \hline
	\end{tabular}
 \vspace{-2em}
\end{table}

\section{Experiments And Results}
\subsection{Experimental Setup}
%\subsubsection{Datasets}
We experiment on two datasets, WEIBO~\cite{jin2017multimodal} and PHEME~\cite{zubiaga2017exploiting}. The Chinese WEIBO dataset consists of fake news from the microblog rumor dispelling system and the real news confirmed by Xinhua News Agency. There are $1.2\%$ news without text and $16.4\%$ news without image. The English PHEME dataset only contains news with missing images.
Moreover, to simulate the harsh practical conditions, we extracted 1000 complete news samples from the WEIBO dataset to create a WEIBO\_Partial dataset. We randomly remove one modality (text or image) from each news sample to make the whole dataset fully contain partial modality. Eventually, the WEIBO\_Partial dataset consists of 500 News with missing texts and 500 news with missing images. We use 800 samples for training and 200 samples for testing.

% To simulate the harsh practical application conditions, we extracted 1000 complete news samples from the WEIBO dataset to create a WEIBO\_Partial dataset, where 800 samples are used for training and 200 samples are used for testing. 
% In the data generation process of WEIBO\_Partial, we randomly remove one modality (text or image) from each news sample to make the whole dataset fully contain partial modality. Eventually, the WEIBO\_Partial dataset consists of 500 News with missing texts and 500 news with missing images.

We compare MUSE with single-modal and multimodal models. Single-modal models consist of GRU~\cite{ma2016detecting}, CAMI~\cite{yu2017convolutional}, %SVM-TS \cite{ma2015detect} 
and TextGCN~\cite{DBLP:conf/aaai/YaoM019}.
%(1) \textbf{GRU} \cite{ma2016detecting} is based on the RNN architecture to capture the hidden representations of tweets over time.
%(2) \textbf{CAMI} \cite{yu2017convolutional} is a CNN based model, which can flexibly extract key features scattered in the input sequence.
%(3) \textbf{SVM-TS} \cite{ma2015detect}  uses time series modeling technology to integrate the social context information of tweets, and uses the linear SVM as the classifier.
%(4) \textbf{TextGCN} \cite{DBLP:conf/aaai/YaoM019} uses graph convolution neural network to learn the feature embeddings of news samples.
% With the development of social media, traditional single modal news has gradually evolved into multimodal news with images or videos. Researchers began to use the deep learning method to learn the multimodal feature of news, which has achieved better fake news detection performance.
Multimodal models include EANN~\cite{wang2018eann}, MVAE~\cite{khattar2019mvae}, %SAFE \cite{zhou2020mathsf}, 
att\_RNN~\cite{jin2017multimodal}, SpotFake*~\cite{singhal2019spotfake}, SpotFake+~\cite{DBLP:conf/aaai/SinghalKSS0K20},  HMCAN~\cite{qian2021hierarchical}, MCAN~\cite{wu2021multimodal}, CAFE~\cite{chen2022cross} and FND-CLIP~\cite{DBLP:conf/icmcs/ZhouYYQZ23}. %, which is introduced as follows.
%(1) \textbf{EANN} \cite{wang2018eann} introduces event classification as an auxiliary task for fake news detection.
%(2) \textbf{MVAE} \cite{khattar2019mvae} uses a bimodal variational autoencoder to obtain the potential representation of news, and classifies the news using a binary classifier.
%(3) \textbf{SAFE} \cite{zhou2020mathsf} incorporates the modal consistency into the fake news detection framework.
%(4) \textbf{att\_RNN} \cite{jin2017multimodal} combines the attention mechanism with RNN to efficiently integrate the multimodal features of news.
%(5) \textbf{SpotFake*} \cite{singhal2019spotfake} uses the pre-trained model to extract news features, and employs the linear layer as the classifier, which is a reproduction of SpotFake \cite{wu2021multimodal}.
%(6) \textbf{SpotFake+} \cite{DBLP:conf/aaai/SinghalKSS0K20} is an improved version of SpotFake, which uses the XLNet model \cite{yang2019xlnet} as the text feature extractor.
%(7) \textbf{HMCAN} \cite{qian2021hierarchical} inputs the multimodal features of samples into the contextual attention network to fuse both inter-modality and intra-modality relationships.
%(8) \textbf{MCAN} \cite{wu2021multimodal} adopts a novel co-attention networks to better fuse textual and visual features for fake news detection.
%(9) \textbf{CAFE} \cite{chen2022cross} is a cross-modal ambiguity learning model, which addresses the misclassified samples by considering the disagreement between different modalities.
% As for the proposed models, 
Moreover, we discretize the search results to obtain the discretized prediction model MUSE-discrete following DARTS. %The discretized model is retained and named as MUSE-discrete in our evaluation. %In order to make fair comparisons across all the used models, we summarized the extracted features and optimization strategies used by each model in Table \ref{tab:models}. We use classification accuracy as the main evaluation metric, and use Precision, Recall, F1 score as complementary evaluation metrics in our experiments.

\subsection{Performance Comparison}
%The results on WEIBO/PHEME are shown in 
Table \ref{tab:comparePHEME} shows that MUSE and its discrete variant MUSE-discrete excel in Accuracy and F1 score on WEIBO, surpassing all the baselines. On PHEME, MUSE achieved comparable performance with HMCAN, possibly due to data quality issues such as label imbalance and sample reuse in PHEME. MUSE's advanced performance stems from its adaptive architecture and efficient noise reduction in incomplete data. 
%WEIBO\_Partial dataset results are in Appendix~\ref{sec:appendix_result}.

% The results on WEIBO/PHEME are shown in Table \ref{tab:comparePHEME}. Based on the results on WEIBO, we observe that MUSE and its discrete version MUSE-discrete have achieved advanced detection performance and outperform all the baseline models in terms of Accuracy and F1 score on two datasets. The results on PHEME show that MUSE achieved comparable performance with HMCAN. This may be because there exist data quality issues such as label imbalance and sample reuse in PHEME, which misguides the models in searching suitable architectures for fake news detection. Overall, MUSE achieves advanced performance that can be attributed to 1) The dynamic paths of MUSE enabling it to adaptively adjust the model architecture according to data characteristics and 2) the incomplete data scheme of MUSE allowing it to de-noise samples efficiently. Furthermore, the results of the WEIBO\_Partial dataset are shown in Appendix~\ref{sec:appendix_result}.

Table \ref{tab:partial_experiment} shows the results on the WEIBO\_Partial dataset. Here we select 1) MCAN \cite{wu2021multimodal}, a model with slightly weaker detection performance than MUSE on the WEIBO dataset, and 2) a simple baseline model that concatenates the text and image modalities using two linear layers as the compared models. The results show that with the rate of partial modality contained news increases (16.4\% to 50\%), the performance difference between these models becomes more obvious. As MCAN focuses on exploiting inter/intra-modal relations, it incorrectly amplifies the noises in the low-quality samples, making the detection performance even 1.2\% lower than a simple baseline model. On the contrary, MUSE exhibits strong adaptability to data quality and outperforms all baseline models in terms of Accuracy and F1 score.

\subsection{Ablation Experiments for Operators}

To figure out whether the number of retained operators during a discrete process affects the detection performance. According to the size of the weights, we remove the operator with the lowest weight, and then retrain the model using the same settings. The results obtained are shown in Table \ref{tab:ablation_oper}.

We can find that reducing the number of retained operators usually increases the model's detection performance. This is because the model needs suitable rather than complex operators. Adding more operators introduces noise into the features when the existing operators are already sufficient for the data.
In experiments, we choose to retain one and two operators for the WEIBO and PHEME datasets, respectively.

\subsection{Ablation Experiments for Paths}
%In order to deeply explore the utility of each path, 
We conduct the ablation experiments on each path: (a) w/o Linear: MUSE with the linear path removed, (b) w/o Sequence: MUSE with the sequence path removed and (c) w/o Auxiliary: MUSE with the auxiliary path removed. Results in Table \ref{tab:ablation_path} and Table \ref{tab:ablation_path_pheme} indicate that removing any path from MUSE will degrade the detection performance, showing the effectiveness of each path.

\section{Conclusion}
In this work, we propose a triple path enhanced neural architecture search model MUSE for fake news detection. MUSE incorporates NAS into fake news detection to provide a solution to dynamically adjusting the model architecture according to data characteristics. 
% Two dynamic paths and one static path are integrated and mutual enhanced in discovering the most effective model architecture. 
The experiments demonstrate the effectiveness of MUSE in detecting fake news.
% over state-of-the-art models. 
In the future, we plan to investigate the application of MUSE in resource-constrained scenarios~\cite{zhang2021survey, zong2021fedcmr}.

\bibliographystyle{abbrvnat}
\bibliography{mylib}

\end{document}